\documentclass[10pt,twocolumn,letterpaper]{article}

\usepackage{cvpr}
\usepackage{times}
\usepackage{epsfig}
\usepackage{graphicx}
\usepackage{subcaption}
\usepackage{amsmath}
\usepackage{amssymb}
\usepackage{bm}
\usepackage[table]{xcolor}
\usepackage{multirow}
\usepackage{booktabs} % for pretty plots
\usepackage[pagebackref=true,breaklinks=true,letterpaper=true,colorlinks,bookmarks=false]{hyperref}

\cvprfinalcopy % *** Uncomment this line for the final submission

\def\cvprPaperID{****} % *** Enter the CVPR Paper ID here

\ifcvprfinal\fi
\begin{document}

\setlength\abovedisplayskip{0.5em}
\setlength\belowdisplayskip{0.5em}

%%%%%%%%% TITLE
\title{Ordinal Depth Supervision for 3D Human Pose Estimation}

\author{Georgios Pavlakos$^1$, Xiaowei Zhou$^2$, Kostas Daniilidis$^1$ \\[0ex]
$^1$ University of Pennsylvania \hspace{1em} $^2$ Zhejiang University\\ 
}

\maketitle
%\thispagestyle{empty}

%%%%%%%%% ABSTRACT
\begin{abstract}

Our ability to train end-to-end systems for 3D human pose estimation from single images is currently constrained by the limited availability of 3D annotations for natural images. Most datasets are captured using Motion Capture (MoCap) systems in a studio setting and it is difficult to reach the variability of 2D human pose datasets, like MPII or LSP. To alleviate the need for accurate 3D ground truth, we propose to use a weaker supervision signal provided by the ordinal depths of human joints. This information can be acquired by human annotators for a wide range of images and poses. We showcase the effectiveness and flexibility of training Convolutional Networks (ConvNets) with these ordinal relations in different settings, always achieving competitive performance with ConvNets trained with accurate 3D joint coordinates. Additionally, to demonstrate the potential of the approach, we augment the popular LSP and MPII datasets with ordinal depth annotations. This extension allows us to present quantitative and qualitative evaluation in non-studio conditions. Simultaneously, these ordinal annotations can be easily incorporated in the training procedure of typical ConvNets for 3D human pose. Through this inclusion we achieve new state-of-the-art performance for the relevant benchmarks and validate the effectiveness of ordinal depth supervision for 3D human pose.

\end{abstract}

%%%%%%%%% BODY TEXT
\section{Introduction}
\begin{figure}[t]
	  \centering
	  \includegraphics[width=1\linewidth,trim={0cm 1cm 2cm 0cm},clip]{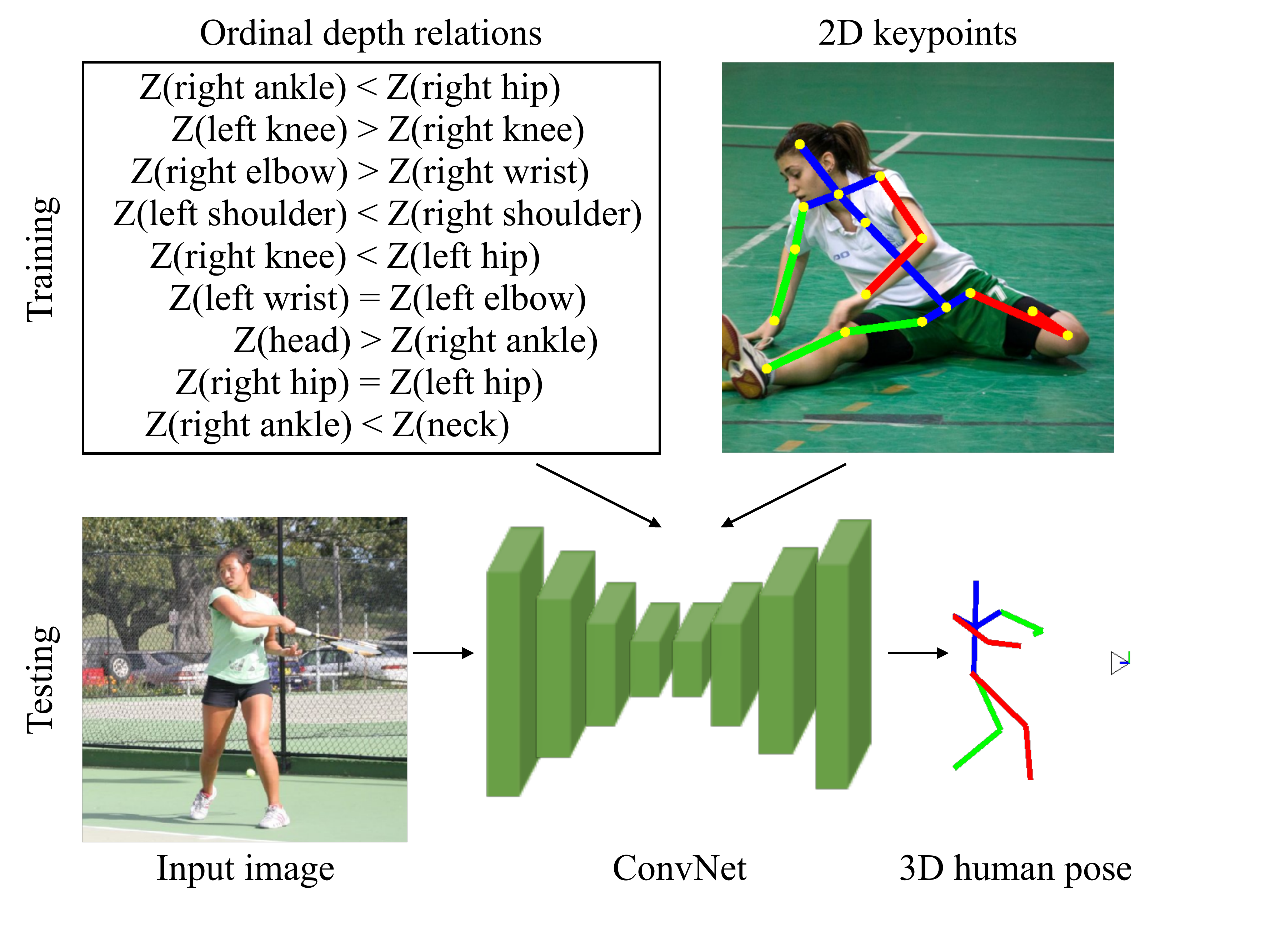} 
    \caption{
Summary of our approach. In the absence of accurate 3D ground truth we propose the use of ordinal depth
relations (closer-farther) of the human body joints for end-to-end training of 3D human pose estimation systems.
}
 \label{fig:pipeline}
\end{figure}

Human pose estimation has been one of the most remarkable successes for deep learning approaches. Leveraging large-scale datasets with extensive 2D annotations has immensely benefited 2D pose estimation~\cite{wei2016cpm,pishchulin2016deepcut,newell2016stacked,yang2017learning}, semantic part labeling~\cite{chen2016attention,xia2016zoom} and multi-person pose estimation~\cite{insafutdinov2016articulated,newell2016associative,cao2016realtime}. In contrast, the complexity of collecting images with corresponding 3D ground truth has constrained 3D human pose datasets in small scale~\cite{kazemi2013multi} or strictly in studio settings~\cite{sigal2010humaneva,ionescu2014human}. The goal of this paper is to demonstrate that in the absence of accurate 3D ground truth, end-to-end learning can be competitive by using weaker supervision in the form of ordinal depth of the joints (Figure~\ref{fig:pipeline}).

Aiming to boost end-to-end discriminative approaches, different techniques attempt to augment the training data. Synthetic examples can be produced in abundance~\cite{chen2016synthesizing,varol2017learning}, but there is no guarantee that they come from the same distribution as natural images. Multi-view systems for accurate capture of 3D ground truth can work outdoors~\cite{mehta2017monocular}, but they need to be synchronized and calibrated, so data collection is not practical and hard to scale. These limitations have favored reconstruction approaches, e.g.,~\cite{bogo2016keep,martinez2017simple}, which employ reliable 2D pose detectors and recover 3D pose in a subsequent step using the 2D joint estimates. Unfortunately, even in the presence of perfect 2D correspondences, the final 3D reconstruction can be erroneous. This 2D-to-3D reconstruction ambiguity is mainly attributed to the binary ordinal depth relations of the joints (closer-farther)~\cite{taylor2000reconstruction}. Leveraging image-based evidences, such as occlusion and shading, can largely resolve the ambiguity, yet this information is discarded by reconstruction approaches.

Motivated by the particular power of ordinal depth relations at resolving reconstruction ambiguities and the fact that this information can be acquired by human annotators, we propose to use ordinal depth relations to train ConvNets for 3D human pose estimation. Since humans can easily perceive pose~\cite{marinoiu2016pictorial} and they are better at estimating ordinal depth than explicit metric depth~\cite{todd2003visual}, annotators can provide pairwise ordinal depth relations for a wide range of imaging conditions, activities, and viewpoints. We develop on the idea of ordinal relations demonstrating their flexibility and effectiveness in a variety of settings: 1) we use them to predict directly the depths of joints, 2) we combine them with 2D keypoint annotations to predict 3D poses, 3) we demonstrate how they can be incorporated within a volumetric representation of 3D pose~\cite{pavlakos2016coarse}.	In every case, the weak supervision signal provided by these ordinal relations leads to a competitive performance compared to fully supervised approaches that employ the actual 3D ground truth. Additionally, to motivate the use of ordinal depth relations for human pose, we provide ordinal depth annotations for two popular 2D human pose datasets, LSP~\cite{johnson2010clustered} and MPII~\cite{andriluka2014mpii}. This extension allows us to provide quantitative and qualitative evaluation of our approach in non-studio settings. Simultaneously, these ordinal annotations for in-the-wild images can be easily incorporated in the training procedure of typical ConvNets for 3D human pose leading to new state-of-the-art results for the standard benchmarks of Human3.6M and HumanEva-I. These performance benefits underline the effectiveness of ordinal depth supervision for human pose problems and provide motivation for further exploration using the available annotations.

Our contributions can be summarized as follows:
\begin{itemize}
\item We propose the use of ordinal depth relations of human joints for 3D human pose estimation to bypass the need for accurate 3D ground truth.
\item We showcase the flexibility of the ordinal relations by incorporating them in different network settings, where we always achieve competitive results to training with the actual 3D ground truth.
\item We augment two popular 2D pose datasets (LSP and MPII) with ordinal depth annotations and demonstrate the applicability of the proposed approach to 3D pose estimation in non-studio conditions.
\item We include our ordinal annotations in the training procedure of typical ConvNets for 3D human pose and exemplify their effectiveness by achieving new state-of-the-art results on the standard benchmarks.
\end{itemize}

%-------------------------------------------------------------------------
\section{Related work}
Since the literature on 3D human pose estimation is vast, here we discuss works closely related to our approach and refer the interested reader to Sarafianos \etal~\cite{sarafianos20163d} for a recent survey on this topic.

\noindent
\textbf{Reconstruction approaches}:
A long line of approaches follows the reconstruction paradigm by employing 2D pose detectors to localize 2D human joints and using these locations to estimate plausible 3D poses~\cite{chen20163d,jahangiri2017generating}. Zhou \etal~\cite{zhou2016sparseness,zhou2017monocap} use 2D heatmaps from a 2D pose ConvNet to reconstruct 3D pose in a video sequence. Bogo \etal~\cite{bogo2016keep} fit a statistical model of 3D human shape to the predicted 2D joints. Alternatively, a network can also handle the step of lifting 2D estimates to 3D poses~\cite{wu2016single,moreno20163d,tome2017lifting}. Notably, Martinez~\etal~\cite{martinez2017simple} achieve state-of-the-art results with a simple multilayer perceptron that regresses 3D joint locations, given 2D keypoints as input. Despite the success of this paradigm, it comes with important drawbacks. No image-based evidence is used during the reconstruction step, the result is too reliant on an imperfect 2D pose detector and even for perfect 2D correspondences, the 3D estimate might fail because of the reconstruction ambiguity. In contrast, by using ordinal depth relations we can leverage rich image-based information during estimation, without relinquishing the accuracy of reconstruction approaches, which can also be integrated in our framework (Section~\ref{sec:reconstruction}).

\noindent
\textbf{Discriminative approaches}:
Discriminative approaches are orthogonal to the reconstruction paradigm since they estimate the 3D pose directly from the image. Prior work uses ConvNets to regress the coordinates of the 3D joints~\cite{li20143d,tekin2016structured,tekin2017learning,tome2017lifting,sun2017compositional,mehta2017monocular}, to regress 3D heatmaps~\cite{pavlakos2016coarse}, or to classify each image in the appropriate pose class~\cite{rogez2016mocap,rogez2017lcr}. The main critique of these end-to-end approaches is that images with corresponding 3D ground truth are required for training. Our work attempts to relax this important constraint, by training with weak 3D information in the form of ordinal depth relations for the joints and 2D keypoints. Weak supervision was also used in recent work~\cite{zhou2017towards} by constraining the lengths of the predicted limbs. However, we argue that our supervision does not simply constraint the output of the network, but also provides novel information for in-the-wild images and further enhances training.

\noindent
\textbf{Generating training examples}:
The limited availability of 3D ground truth for training 3D human pose ConvNets has also been addressed in various ways in recent works. The most straightforward solution is to use graphics to augment the training data~\cite{chen2016synthesizing,varol2017learning,mehta2017monocular}. Differently, Rogez and Schmid~\cite{rogez2016mocap} propose a collage approach by composing human parts from different images to produce combinations with known 3D pose. In both cases though, most examples do not reach the detail and variety level that in-the-wild images have. Mehta \etal~\cite{mehta2017monocular} record multiple views outdoors and estimate accurate 3D ground truth for every view. However, multi-view systems need to be synchronized and calibrated, so large-scale data collection is not trivial.

\noindent
\textbf{3D annotations}:
Prior works have also relied on humans to perceive and annotate 3D properties that are lost through the projection of a 3D scene on a 2D image. Bell~\etal~\cite{bell2014intrinsic} and Chen \etal~\cite{chen2016single} annotate the ordinal relations for the apparent depth of pixels in the image. In the work of Xiang~\etal~\cite{xiang2014beyond,xiang2016objectnet3d}, humans align 3D CAD models with single images to provide viewpoint information. Concerning 3D human pose annotations, the famous poselets work from Bourdev and Malik~\cite{bourdev2009poselets} uses an interactive tool for annotators to adjust the 3D pose, making the procedure laborious. Maji~\etal~\cite{maji2011action} provide 3D annotations for human pose, but only in the form of yaw angles for head and torso. The idea of ordinal depth relations is also explored by Pons-Moll \etal~\cite{pons2014posebits} where attributes regarding the relative 3D position of the body parts are included in their posebits database. Different to them, we provide annotations by humans for a much larger set of images (i.e., more  than 15k images with our annotations compared to 1k for the posebits dataset), and instead of exploring an extensive set of pose attributes, we propose a cleaner training scheme that requires only 2D keypoint locations and ordinal depth relations. In recent work, Lassner~\etal~\cite{lassner2017unite} estimate proposals of 3D human shape fits for single images which are accepted or rejected by annotators. Despite the rich ground truth in case of a good fit, many automatic proposals are of low quality, leading to many discards. Our work aims for a more balanced solution where 3D annotations have a weaker form, but the task is easy for humans, so that they can provide annotations on a large scale for practically any available image.

\noindent
\textbf{Ordinal relations}:
There is a long history for learning from ordinal relations, outside the field of computer vision, with particular interest in the area of information retrieval, where many algorithms for learning-to-rank have been developed~\cite{burges2005learning,cao2007learning,taylor2008softrank}. In the context of computer vision, previous works have used relations to learn apparent depth~\cite{zoran2015learning,chen2016single} or reflectance~\cite{narihira2015learning,zhou2015learning} of a scene. We share a common motivation with these approaches in the sense that ordinal relations are easier for humans to annotate, compared to metric depth or absolute reflectance values.

\section{Technical approach}\label{sec:approach}
In this section we present our proposed approach for different settings of 3D human pose estimation. First, in Section~\ref{sec:depthPred} we predict only the depths of the human joints, relying on ordinal depth relations and a ranking loss for training. Then, in Section~\ref{sec:coordPred} we combine the ordinal relations with 2D keypoint annotations to predict the 3D pose coordinates. In Section~\ref{sec:volumePred} we explore the incorporation of ordinal relations within a volumetric representation for 3D human pose~\cite{pavlakos2016coarse}. Finally, Section~\ref{sec:reconstruction} presents the extension of the previous
networks with a component designed to encode a geometric 3D pose prior.

\subsection{Depth prediction}\label{sec:depthPred}
Our initial goal is to establish the training procedure such that we can leverage ordinal depth relations to learn to predict the depths of human joints. This is the simplest case, where instead of explicitly predicting the 3D pose, we only predict depth values for the joints.

Let us represent the human body with $N$ joints. For each joint $i$ we want to predict its depth $z_i$. The provided data are in the form of pairwise ordinal depth relations. For a pair of joints $(i,j)$, we denote the
ordinal depth relation as $r_{(i,j)}$ taking the value:
\begin{itemize}
\item $+1$, if joint $i$ is closer than $j$,
\item $-1$, if joint $j$ is closer than $i$,
\item $0$, if their depths are roughly the same.
\end{itemize}
The ConvNet we use for this task takes the image as input and predicts $N$ depth values $z_i$, one for each joint. Given the $r_{(i,j)}$ relation and assuming that the ConvNet is producing the depth estimates $z_i$ and $z_j$ for the two corresponding joints, the loss for this pair is:
\begin{eqnarray}
\mathcal{L}_{i,j} =\left\{
\begin{array}{ll}
\log\left(1+\exp(z_i - z_j) \right), & r_{(i,j)} = +1 \\
\log\left(1+\exp(-z_i + z_j) \right),  &  r_{(i,j)} = -1 \\
 (z_i - z_j)^2, &  r_{(i,j)} = 0. \\
\end{array}
\right.
\label{equation:ranking_loss}
\end{eqnarray}
This is a differentiable ranking loss expression, which has similarities with early works on the learning-to-rank literature~\cite{burges2005learning} and was also adopted by~\cite{chen2016single} for apparent depth estimation. Intuitively, it enforces a large margin between the values $z_i$ and $z_j$ if one of them has been annotated as closer than the other, otherwise it enforces them to be equal. Denoting with $\mathcal{I}$ the set of pairs of joints that have been annotated with an ordinal relation, the complete expression for the loss takes the form: 
\begin{eqnarray}
\mathcal{L}_{rank} = \sum_{(i,j) \in \mathcal{I}} \mathcal{L}_{i,j}.
\label{equation:full_ranking_loss}
\end{eqnarray}

An interesting property of this loss is that we do not require the relations for all pairs of joints to be available during training. The loss can be computed based only on the subset of pairs that have been annotated. Additionally, the relations do not have to be consistent, i.e., no strict global ordering is required. Instead, the ConvNet is allowed to learn a consensus from the provided relationships by minimizing the incurred loss. This is a helpful property in case there are ambiguities in the annotations.

\subsection{Coordinate prediction for 3D pose}\label{sec:coordPred}
Our initial ConvNet only predicts the depths of the human joints. To enable full 3D pose reconstruction, we additionally need to precisely localize the corresponding joints on the image. Given the ConvNet used in the previous section, the most natural extension is to enrich its output by predicting the 2D coordinates of the joints as well. Thus, we predict $2N$ additional values which correspond to the pixel coordinates $\bm{w} = (x,y)$ of each joint. We consider this combination of 2D keypoints with ordinal depth as a form of \emph{weak 3D information} and we refer to the corresponding ConvNet as the \emph{weakly supervised version}.

Let us denote with $\bm{w}_n$ the ground truth 2D location for joint $n$, and with $\hat{\bm{w}}_n$ the corresponding ConvNet prediction. Assuming the availability of 2D keypoint annotations, the familiar $\mathcal{L}_2$ regression loss can be applied:
\begin{eqnarray}
\mathcal{L}_{keyp} = \sum_{n=1}^N \| \bm{w}_n - \hat{\bm{w}}_n \|_2^2.
\end{eqnarray}
By combining the ranking loss for the values $z_n$ and the regression loss for the keypoint coordinates $\bm{w}_n$, we can train the ConvNet end-to-end: $\mathcal{L} = \mathcal{L}_{rank} + \lambda \mathcal{L}_{keyp}$, where the value $\lambda = 100$ is used for our experiments. 

\subsection{Volumetric prediction for 3D pose}\label{sec:volumePred}
Apart from direct regression of the 3D pose coordinates, recent work has investigated the use of a volumetric representation for 3D human pose~\cite{pavlakos2016coarse}. In this case, the space around the subject is discretized, and the ConvNet predicts per-voxel likelihoods for every joint in the 3D space. The training target for the volumetric space is a 3D Gaussian centered at the 3D location of each joint. However, without explicit 3D ground truth, supervising the same volume is not trivial. To demonstrate the general applicability of ordinal relations, we adapt this representation, to make it compatible with ordinal depth supervision as well.

\begin{figure}[t]
	  \centering
	  \includegraphics[width=1\linewidth,trim={1cm 0cm 2cm 0cm},clip]{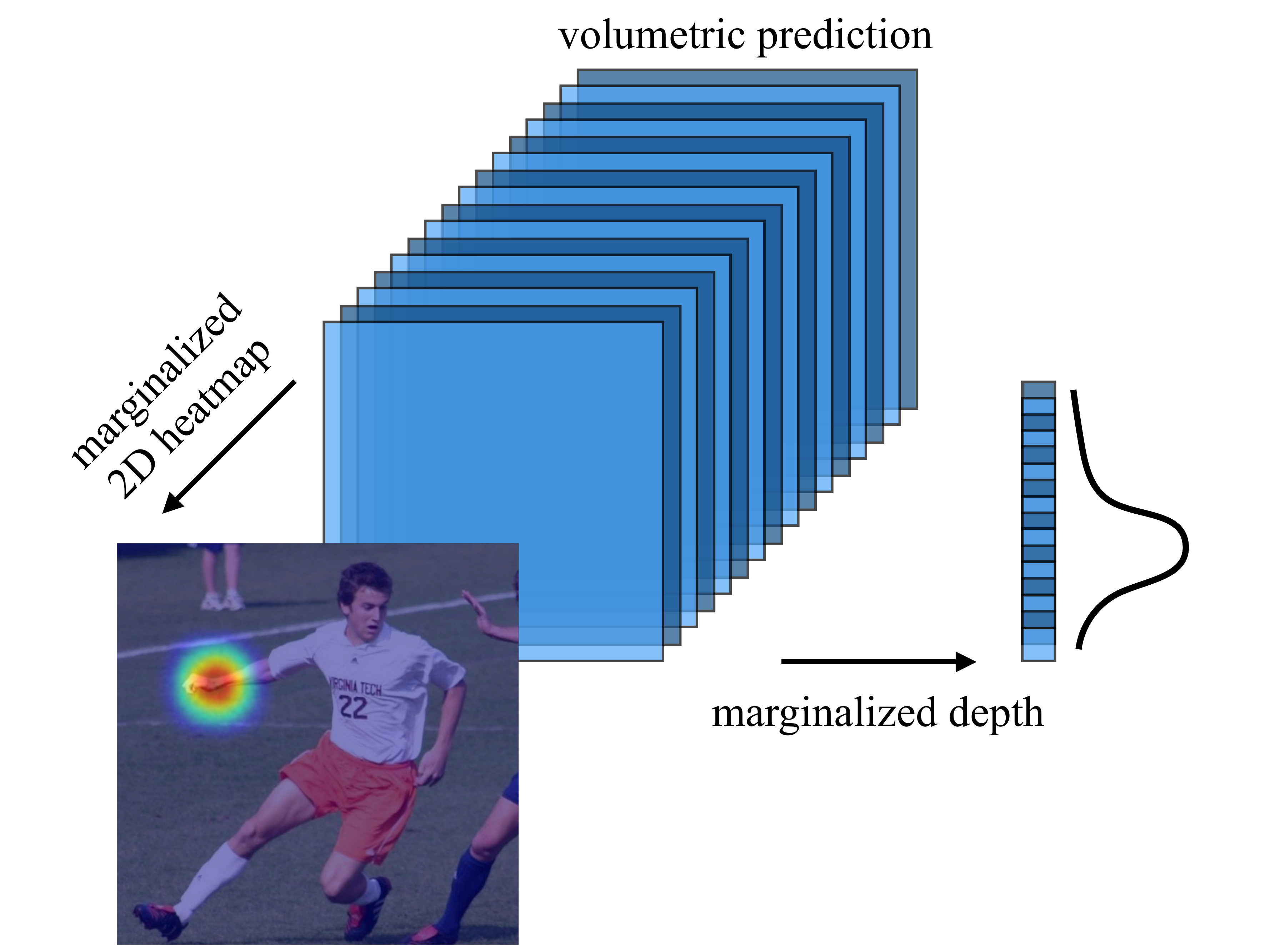}
    \caption{
Visualization of the volumetric output for an individual joint. The predictions are volumetric, but in the absence of accurate 3D ground truth, the supervision is applied independently on the 2D image plane and the depth dimension. The marginalized likelihoods are computed by means of sum-pooling operations.
}
 \label{fig:marginalized}
\end{figure}

To bypass the seemingly complex issue, we propose to preserve the volumetric structure of the output, but decompose the supervision a) in the 2D image plane and b) the $z$ dimension (depth), as presented in Figure~\ref{fig:marginalized}. Precisely, for every joint $n$, the ConvNet predicts score maps $\Psi_n$, which can be transformed to a probability distribution, by applying a {\em softmax} operation $\sigma$. So, the joint $n$ is located in position $\bm{u} = (x,y,z)$ with probability $p(\bm{u}|n) = \sigma[\Psi_n]_{\bm{u}}$. The marginalized probability distribution in the 2D plane is:
\begin{eqnarray}
\label{eq:marg_2D}
p(x,y|n) = \sum_z p(\bm{u}|n),
\end{eqnarray}
and can be computed efficiently as a sum-pooling operation across all the slices of the volume. This operation is equivalent to adopting a weak perspective camera model. Similarly, the marginalized probability distribution for the depth dimension is:
\begin{eqnarray}
\label{eq:marg_depth}
p(z|n) = \sum_{x,y} p(\bm{u}|n),
\end{eqnarray}
and can again be computed as a sum-pooling operation across all the pixels of a slice. This decomposition has the advantage that even if we do not have complete 3D ground truth, we can still supervise the ConvNet. The 2D image plane (values of equation~\ref{eq:marg_2D}) and the depth dimension (values of equation~\ref{eq:marg_depth}) are supervised independently, but they are connected by the underlying volumetric representation which enforces the 3D consistency. Our loss function takes the form: $\mathcal{L} = \mathcal{L}_{rank} + \lambda \mathcal{L}_{heat}$. The loss for the $z$-dimension, $\mathcal{L}_{rank}$, is the same ranking loss as before (equation~\ref{equation:full_ranking_loss}), where we recover depth for each joint by taking the mean value of the estimated soft distribution: $z_n = \sum_{z}z p(z|n)$. For the $x$-$y$ dimensions, the target for each keypoint is a heatmap with a Gaussian centered around its ground truth location and $\mathcal{L}_{heat}$ is an $\mathcal{L}_2$ loss between the predicted and the ground truth heatmaps~\cite{tompson2014joint,pfister2015flowing}.

We stress here that the alterations presented up to this point refer only to the supervision type, without interfering with the network architecture. This allows most of the state-of-the-art discriminative ConvNets~\cite{zhou2017towards,sun2017compositional,tekin2017learning,pavlakos2016coarse} to be used as-is, and be complemented with the proposed ordinal depth supervision when 3D ground truth is not available.
\subsection{Integration with a reconstruction component}\label{sec:reconstruction}
\begin{figure}
    \centering
    \begin{subfigure}[b]{0.45\textwidth}
        \includegraphics[width=1\linewidth,trim={0cm 6cm 0cm 5cm},clip]{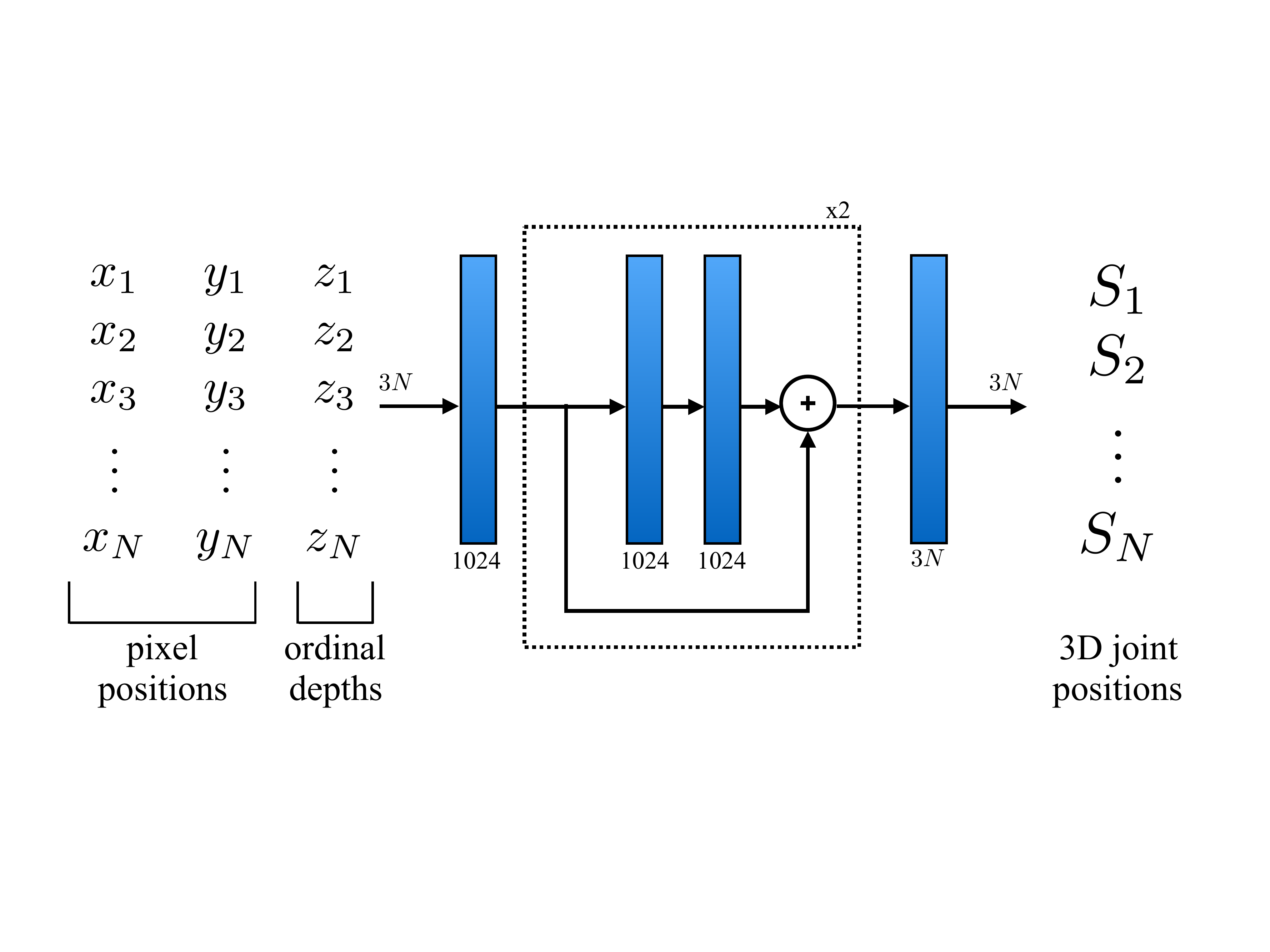}
        \caption{The reconstruction component.}
        \label{fig:reconstruction}
    \end{subfigure}
    ~
    \begin{subfigure}[b]{0.45\textwidth}
        \includegraphics[width=1\linewidth,trim={0cm 22cm 15cm 0cm},clip]{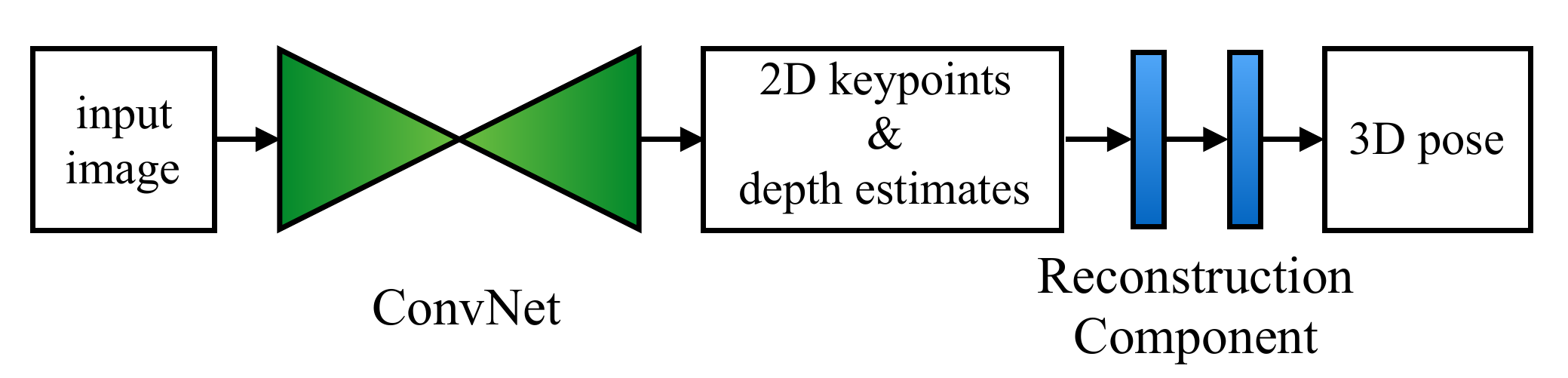}
        \caption{Integration of the reconstruction component.}
        \label{fig:everything}
    \end{subfigure}
    ~
    \caption{
(a) The reconstruction component is a multi-layer perceptron with two bilinear units~\cite{martinez2017simple}. The input is the concatenation of the pixel locations of the joints $(x_i,y_i)$, and the ordinal depths $z_i$, while the output is the 3D pose coordinates $S_i$.
(b) Integration of the reconstruction module in the full framework. The ConvNet of Section~\ref{sec:coordPred} or~\ref{sec:volumePred} estimates 2D keypoint locations and depths which are used by the reconstruction module to predict a coherent 3D pose.
}\label{fig:priors}
\end{figure}

The strength of the aforementioned networks is that they leverage image-based information to resolve the single-view depth ambiguities and produce depth estimates $z_n$ that respect the ordinal depths of the human joints. However, the predicted depth values do not typically match the exact metric depths of the joints, since no full 3D pose example has been used to train the networks. This motivates us to enhance the architecture with our proposed {\em reconstruction} component, which takes as input the estimated 2D keypoints $\bm{w}_n$ and the ordinal depth estimates $z_n$, for all joints $n$, and reconstructs the 3D pose, $S \in \mathbb{R}^{n \times 3}$. This input-output relation is presented in Figure~\ref{fig:reconstruction}. Conveniently, for the training of this component we require only MoCap data, which are available in abundance. During training, we simply project each 3D pose skeleton to the 2D image plane. To simulate the input, we use the projected 2D joint locations and a noisy version of the depths of the joints, such that the majority of their ordinal relations are preserved, while their values might not necessarily match the actual depth. Denoting with $\hat{S}_i$ the output 3D joints of the ConvNet and with $S_i$ the joints of the 3D pose that was used to generate the input, our supervision is an $\mathcal{L}_2$ loss:
\begin{eqnarray}
\mathcal{L}_{3D} = \sum_{n=1}^N \| S_n - \hat{S}_n \|_2^2.
\end{eqnarray}
This module can be easily incorporated in an end-to-end framework by using as input the output of the ConvNet from Section~\ref{sec:coordPred} or Section~\ref{sec:volumePred}. This is presented schematically in Figure~\ref{fig:everything}. The benefit from employing such a reconstruction module
is demonstrated empirically in Section~\ref{sec:experiments}.

\section{Empirical evaluation}\label{sec:experiments}
This section concerns the empirical evaluation of the proposed approach. First, we present the benchmarks that we employed for quantitative and qualitative evaluation. Then, we provide some essential implementation details of the approach. Finally, quantitative and qualitative results are presented on the selected datasets.

\subsection{Datasets}

We employed two standard indoor benchmarks, Human3.6M~\cite{ionescu2014human} and HumanEva-I~\cite{sigal2010humaneva}, along with a recent dataset captured in indoor and outdoor conditions, MPI-INF-3DHP~\cite{mehta2017monocular,mehta2017vnect}. Additionally, we extended two popular 2D human pose datasets, Leeds Sports Pose dataset (LSP)~\cite{johnson2010clustered} and MPII human pose dataset (MPII)~\cite{andriluka2014mpii} with ordinal depth annotations for the human joints.

\noindent
\textbf{Human3.6M}: It is a large-scale dataset captured in an indoor environment that contains multiple subjects performing typical actions like ``Eating'' and ``Walking''. Following the most popular protocol (e.g.,~\cite{zhou2016sparseness}), we train using subjects S1,S5,S6,S7, and S8 and test on subjects S9 and S11. The original videos are downsampled from 50fps to 10fps to remove redundancy. A single model is trained for all actions. Results are reported using the mean per joint error and the reconstruction error, which allows a Procrustes alignment of the prediction with the ground truth.

\noindent
\textbf{HumanEva-I}: It is a smaller scale dataset compared to Human3.6M, including fewer users and actions. We follow the typical protocol (e.g.,~\cite{bo2010twin}), where the training sequences of subjects S1, S2 and S3 are used for training and the validation sequences of the same subjects are used for testing. We train a single model for all actions and users, and we report results using the reconstruction error.

\noindent
\textbf{MPI-INF-3DHP}: 
It is a recent dataset that includes both indoor and outdoor scenes. We use it exclusively for evaluation, without employing the training data, to demonstrate robustness of the trained model under significant domain shift. Following the typical protocol (\cite{mehta2017monocular,zhou2017towards}), results are reported using the PCK3D and the AUC metric.

\noindent
\textbf{LSP + MPII Ordinal}:
Leeds Sports Pose~\cite{johnson2010clustered} and MPII human pose~\cite{andriluka2014mpii} are two of the most widely used benchmarks for 2D human pose. Here we extend both of them, offering ordinal depth annotations for the human joints. For LSP we annotate all the 2k images, while for MPII we annotate the subset of 13k images used by Lassner~\etal~\cite{lassner2017unite}.

Annotators were presented with a pair of joints for each image and answered which joint was closer to the camera. The option ``ambiguous/hard to tell'' was also offered. We considered 14 joints, excluding thorax and spine joints of MPII, which are often not used for training (e.g.,~\cite{wei2016cpm}). The questions for each image were continued until a global ordering could be inferred for all the joints. By enforcing a global ordering we conveniently do not encounter any contradicting annotations. More importantly though, this approach significantly decreased annotation time. If the relative questions had to be answered for all joints, then we would require ${14 \choose 2} = 91$ questions for each image. In contrast, with the procedure we followed, we could get a global ordering with roughly 17 questions per image in the mean case. This resulted in 5 times faster annotation time. Additionally, we observed that annotators were much more efficient when they were asked continuously about a specific pair of joints, instead of changing the pair of focus. As a result, we created groups of 50 images containing questions about the same pair of joints. This way we could get annotations at a rate of 3.5 secs per question, meaning that in total the procedure required roughly 1 minute per image.

We clarify that our goal for this dataset is to provide a novel information source (ordinal depth) for in-the-wild images. We do not use it for evaluation, since it is not a mm level accuracy benchmark like Human3.6M or HumanEva-I. Furthermore, the goal is not to conduct a computational study concerning the level of accuracy that humans perceive 3D poses as this has been already examined in the past~\cite{marinoiu2016pictorial}. In contrast, we use these annotations to demonstrate that: a) they can boost performance of 3D human pose estimation for standard benchmarks, and b) they assist our ConvNets to proper generalize and make them applicable in non-studio conditions, or in cases with significant domain shift.

\begin{table}
\centering
\small
\hspace{-3mm}
\tabcolsep=1.0mm
\begin{tabular}{@{} cccc @{}}
\toprule
\multicolumn{2}{c}{Architecture} & Supervision & Avg error \\
\midrule
\multicolumn{2}{c}{depth} & ordinal supervision & 84.24 \\
\multicolumn{2}{c}{prediction} & direct regression & 80.23  \\
\midrule
\multicolumn{2}{c}{coordinate} & weakly supervised & 115.08 \\
\multicolumn{2}{c}{regression} & fully supervised~\cite{pavlakos2016coarse} & 112.41 \\ 
\midrule
 & one & weakly supervised & 89.93 \\
\multirow{2}{*}{volume} & hourglass & fully supervised~\cite{pavlakos2016coarse} & 85.82 \\
\cmidrule{2-4}
regression & two &  weakly supervised & 79.03 \\
 & hourglasses & fully supervised~\cite{pavlakos2016coarse} & 69.77 \\
\bottomrule
\end{tabular}
\caption{Effect of training with the actual 3D ground truth, versus employing weaker ordinal depth supervision on Human3.6M. The results are mean per joint errors (mm).
}
\label{tab:hm36m_ordinal}
\end{table}

\subsection{Implementation details}\label{sec:details}
For the ConvNets that predict 2D keypoints and/or depths, we follow the hourglass design~\cite{newell2016stacked}. When the output is in coordinate form (Sections~\ref{sec:depthPred} and~\ref{sec:coordPred}), we use one hourglass with a fully connected layer in the end, while when we have volumetric target (Section~\ref{sec:volumePred}), we use two hourglasses (unless stated otherwise). For comparisons with the state-of-the-art, we follow a mixed training strategy combining images with 3D ground truth from the respective dataset (Human3.6M or HumanEva-I), with LSP+MPII Ordinal images. For the LSP+MPII Ordinal examples, the loss is computed based on the human annotations (weak supervision), while for the respective dataset examples, the loss is computed based on the known ground truth (full supervision). We train the network with a batch size of 4, learning rate set to 2.5e-4, and using rmsprop for the optimization. Augmentation for rotation ($\pm 30^{\circ}$), scale (0.75-1.25) and flipping (left-right) is also used. The duration of the training depends on the size of the dataset (300k iterations for Human3.6M data only, 2.5M iterations for mixed Human3.6M and LSP+MPII Ordinal data, 1.5M iterations for mixed HumanEva-I and LSP+MPII Ordinal data). For the reconstruction component (Section~\ref{sec:reconstruction}), we follow the design of~\cite{martinez2017simple}. We train the network with a batch size of 64, learning rate set to 2.5e-4, we use rmsprop for the optimization, and the training lasts for 200k iterations.

\begin{table}
\centering
\hspace{-3mm}
\tabcolsep=0.75mm
\begin{tabular}{@{}ccccc@{}}
\toprule
& & & & Avg error \\
\midrule
Human3.6M & & & & 71.9 \\
\midrule
Human3.6M & + 2D keyp & & & 66.6 \\
Human3.6M & + 2D keyp & + Ord & & 62.1 \\
\midrule
Human3.6M & + 2D keyp & & + Rec & 59.1 \\
Human3.6M & + 2D keyp & + Ord & + Rec & {\bf 56.2} \\
\bottomrule
\end{tabular}
\caption{Ablative study on Human3.6M demonstrating the effect of incorporating
additional data sources in the training procedure (2D keypoints and ordinal depth relations),
as well as integrating a rconstruction component. The numbers are mean per joint errors (mm).}
\label{tab:hm36m_ablative}
\end{table}

\begin{table}
\centering
\hspace{-3mm}
\tabcolsep=0.75mm
\begin{tabular}{@{}lcc@{}}
\toprule
 & PCK3D & AUC \\
\midrule
Human3.6M & 17.1 & 6.3 \\
Human3.6M + 2D keyp & 44.3 & 19.8 \\
\midrule
Human3.6M + 2D keyp + Ord  & {\bf 71.9} & {\bf 35.3} \\
\bottomrule
\end{tabular}
\caption{Ablative study on MPI-INF-3DHP demonstrating that supervision through our ordinal annotations is important for proper generalization.}
\label{tab:robustness}
\end{table}

\begin{table*}
\centering
\footnotesize
\hspace{-3mm}
\tabcolsep=0.75mm
\begin{tabular}{@{}lrrrrrrrrrrrrrrrr@{}}
\toprule
 & Direct. & Discuss & Eating & Greet & Phone & Photo & Pose & Purch. & Sitting & SitingD & Smoke & Wait & WalkD & Walk & WalkT & Avg\\
\midrule
Tekin~\etal~\cite{tekin2015direct} (CVPR'16)    & 102.4 & 147.2 & 88.8 & 125.3 & 118.0 & 182.7 & 112.4 & 129.2 & 138.9 & 224.9 & 118.4 & 138.8 & 126.3 & 55.1 & 65.8 & 125.0\\
Zhou~\etal~\cite{zhou2016sparseness} (CVPR'16)    & 87.4 & 109.3 & 87.1 & 103.2 & 116.2 & 143.3 & 106.9 & 99.8 & 124.5 & 199.2 & 107.4 & 118.1 & 114.2 & 79.4 & 97.7 & 113.0\\
Du~\etal~\cite{du2016marker} (ECCV'16)    & 85.1 & 112.7 & 104.9 & 122.1 & 139.1 & 135.9 & 105.9 & 166.2 & 117.5 & 226.9 & 120.0 & 117.7 & 137.4 & 99.3 & 106.5 & 126.5\\
Zhou~\etal~\cite{zhou2016deep} (ECCVW'16)    & 91.8 & 102.4 & 96.7 & 98.8 & 113.4 & 125.2 & 90.0 & 93.8 & 132.2 & 159.0 & 107.0 & 94.4 & 126.0 & 79.0 & 99.0 & 107.3\\
Chen~\etal~\cite{chen20163d} (CVPR'17) & 89.9 & 97.6 & 90.0 & 107.9 & 107.3 & 139.2 & 93.6 & 136.1 & 133.1 & 240.1 & 106.7 & 106.2 & 114.1 & 87.0 & 90.6 & 114.2\\
Tome~\etal~\cite{tome2017lifting} (CVPR'17) & 65.0 & 73.5 & 76.8 & 86.4 & 86.3 & 110.7 & 68.9 & 74.8 & 110.2 & 173.9 & 85.0 & 85.8 & 86.3 & 71.4 & 73.1 & 88.4\\
Rogez~\etal~\cite{rogez2017lcr} (CVPR'17) & 76.2 & 80.2 & 75.8 & 83.3 & 92.2 & 105.7 & 79.0 & 71.7 & 105.9 & 127.1 & 88.0 & 83.7 & 86.6 & 64.9 & 84.0 & 87.7 \\
Pavlakos~\etal~\cite{pavlakos2016coarse} (CVPR'17) & 67.4 & 71.9 & 66.7 & 69.1 & 72.0 & 77.0 & 65.0 & 68.3 & 83.7 & 96.5 & 71.7 & 65.8 & 74.9 & 59.1 & 63.2 & 71.9\\
Nie~\etal~\cite{xiaohan2017monocular} (ICCV'17)  & 90.1 & 88.2 & 85.7 & 95.6 & 103.9 & 103.0 & 92.4 & 90.4 & 117.9 & 136.4 & 98.5 & 94.4 & 90.6 & 86.0 & 89.5 & 97.5\\
Tekin~\etal~\cite{tekin2017learning} (ICCV'17)  & 54.2&  61.4&	60.2&	61.2&	79.4&	78.3&	63.1&	81.6&	70.1&	107.3&	69.3&	70.3&	74.3&	51.8&	74.3&	69.7\\
Zhou~\etal~\cite{zhou2017towards} (ICCV'17)  & 54.8&  60.7&	58.2&	71.4&	62.0&	65.5&	53.8&	55.6&	75.2&	111.6&	64.2&	66.1&	51.4&	63.2&	55.3&	64.9\\
Martinez~\etal~\cite{martinez2017simple} (ICCV'17) & 51.8&  56.2&	58.1&	59.0&	69.5&	78.4&	55.2&	58.1&	74.0&	94.6&	62.3&	59.1&	65.1&	49.5&	52.4&	62.9\\
\midrule
Ours  & \bf{48.5}& 	\bf{54.4}& 	\bf{54.4}& 	\bf{52.0}& 	\bf{59.4}& 	\bf{65.3}& 	\bf{49.9}& 	\bf{52.9}& 	\bf{65.8}& 	\bf{71.1}& 	\bf{56.6}& 	\bf{52.9}& 	\bf{60.9}& 	\bf{44.7}& 	\bf{47.8}& 	\bf{56.2}\\
\bottomrule
\end{tabular}
\vspace{-5pt}
\caption{Detailed results on Human3.6M~\cite{ionescu2014human}. Numbers are mean per joint errors (mm). The results of all approaches are obtained from the original papers. We outperform all other approaches across the table.}
\label{tab:hm36m_mpjpe}
\end{table*}

\begin{table*}
\centering
\footnotesize
\hspace{-3mm}
\tabcolsep=0.6mm
\begin{tabular}{@{}lrrrrrrrrrrrrrrrr@{}}
\toprule
 & Direct. & Discuss & Eating & Greet & Phone & Photo & Pose & Purch. & Sitting & SitingD & Smoke & Wait & WalkD & Walk & WalkT & Avg\\
\midrule
Akhter \& Black~\cite{akhter2015pose}* (CVPR'15) & 199.2 & 177.6 & 161.8 & 197.8 & 176.2 & 186.5 & 195.4 & 167.3 & 160.7 & 173.7 & 177.8 & 181.9 & 176.2 & 198.6 & 192.7 & 181.1\\
Ramakrishna~\etal~\cite{ramakrishna2012}* (ECCV'12) & 137.4 & 149.3 & 141.6 & 154.3 & 157.7 & 158.9 & 141.8 & 158.1 & 168.6 & 175.6 & 160.4 & 161.7 & 150.0 & 174.8 & 150.2 & 157.3\\
Zhou~\etal~\cite{zhou2016sparse}* (CVPR'15) & 99.7 & 95.8 & 87.9 & 116.8 & 108.3 & 107.3 & 93.5 & 95.3 & 109.1 & 137.5 & 106.0 & 102.2 & 106.5 & 110.4 & 115.2 & 106.7\\
Bogo~\etal~\cite{bogo2016keep} (ECCV'16) & 62.0 & 60.2 & 67.8 & 76.5 & 92.1 & 77.0 & 73.0 & 75.3 & 100.3 & 137.3 & 83.4 & 77.3 & 86.8 & 79.7 & 87.7 & 82.3\\
Moreno-Noguer~\cite{moreno20163d} (CVPR'17) & 66.1 & 61.7 & 84.5 & 73.7 & 65.2 & 67.2 & 60.9 & 67.3 & 103.5 & 74.6 & 92.6 & 69.6 & 71.5 & 78.0 & 73.2 & 74.0\\
Pavlakos~\etal~\cite{pavlakos2016coarse} (CVPR'17) & 47.5 & 50.5 & 48.3 & 49.3 & 50.7 & 55.2 & 46.1 & 48.0 & 61.1 & 78.1 & 51.1 & 48.3 & 52.9 & 41.5 & 46.4 & 51.9 \\
Martinez~\etal~\cite{martinez2017simple} (ICCV'17) & 39.5 & 43.2&	46.4&	47.0&	51.0&	56.0&	41.4&	40.6&	56.5&	69.4&	49.2&	45.0&	49.5&	38.0&	43.1&	47.7\\
\midrule
Ours   & \bf{34.7} & \bf{39.8}&	\bf{41.8}&	\bf{38.6}&	\bf{42.5}&	\bf{47.5}&	\bf{38.0}&	\bf{36.6}&	\bf{50.7}&	\bf{56.8}&	\bf{42.6}&	\bf{39.6}&	\bf{43.9}&	\bf{32.1}&	\bf{36.5}&	\bf{41.8}\\
\bottomrule
\end{tabular}
\vspace{-5pt}
\caption{Detailed results on Human3.6M~\cite{ionescu2014human}. Numbers are reconstruction errors. The results of all approaches are obtained from the original papers, except for (*), which were obtained from~\cite{bogo2016keep}. We outperform all other approaches across the table.}
\label{tab:hm36m_rec}
\end{table*}

\subsection{Ablative studies}

\noindent
\textbf{Ordinal supervision}:
First, we examine the effect of using ordinal depth supervision versus employing the actual 3D groudtruth for training. For this part, we focus on Human3.6M which is a large scale benchmark and provides 3D ground truth to perform the quantitative comparison. To define the ordinal depth relations, the depth values for each pair of joints are considered. If they differ less than 100mm, then the corresponding relation is set to $r = 0$ (similar depth). Otherwise, it is set to $r = \pm 1$, depending on which joint is closer. Since for this comparison we want to focus on the form of supervision, this is the only set of experiments that uses ordinal depth relations inferred from 3D ground truth. For the remaining evaluations, all ordinal depth relations were provided by human annotators.

Following the analysis of Section~\ref{sec:approach}, we explore three different prediction schemes, i.e., depth prediction, coordinate regression and volume regression. For each one of them, we compare a version where ordinal supervision is used, versus employing the actual 3D ground truth for training. The detailed results are presented in Table~\ref{tab:hm36m_ordinal}. Interestingly, in all cases, the weaker ordinal supervision signal is competitive and achieves results very close to the fully supervised baseline. The gap increases only when we employ more powerful architectures, i.e., the volume regression case with two hourglass components. In fact, in this case the average error is already very low (below 80mm), and one would expect that for even lower prediction errors, the highly accurate 3D ground truth would be necessary for training.

\noindent
\textbf{Improving 3D pose detectors}:
After the sanity check that ordinal supervision is competitive to training with the full 3D ground truth, we explore using ordinal depth annotations provided by humans, to boost the performance of a standard ConvNet for 3D human pose~\cite{pavlakos2016coarse}. As detailed in Section~\ref{sec:details}, we follow a mixed training strategy, leveraging Human3.6M images with 3D ground truth and LSP+MPII Ordinal images with our annotations. Data augmentation using natural images with 2D keypoint annotations is a standard practice~\cite{tekin2017learning,mehta2017monocular,popa2017deep,zhou2017towards,sun2017compositional}, but here we also consider the effect of our ordinal depth supervision. Optionally, the reconstruction component can be used at the end of the network, helping with coherent 3D pose prediction. The detailed results of the ablative study are presented in Table~\ref{tab:hm36m_ablative}.

Unsurprisingly, using more training examples improves performance. The supervision with 2D keypoints is helpful (line 2), however the addition of our ordinal depth supervision provides novel information to the network and further improves the results (line 3). The refinement step using the reconstruction module (lines 4 and 5) is also beneficial, and helps providing coherent 3D pose results. In fact, the last line corresponds to state-of-the-art results for this dataset, which we discuss in more detail in Section~\ref{sec:sota}.

\noindent
\textbf{Robustness to domain shift}:
Besides boosting current state-of-the-art models, we ultimately aspire to use our ordinal supervision for better generalization of the trained models so that they are applicable for in-the-wild images. To demonstrate this potential, we test our approach on the MPI-INF-3DHP dataset. This dataset is not considered exactly in-the-wild, but has a significant domain shift compared to Human3.6M. The complete results for this ablative experiment are presented in Table~\ref{tab:robustness}. Interestingly, the model trained only on Human3.6M data (line 1) has embarrassing performance, because of heavy overfitting. Using additional in-the-wild images with 2D keypoints (line~2) is helpful, but from inspection of the results, the benefit comes mainly from better 2D pose estimates, while depth prediction is generally mediocre. The best generalization comes after incorporating also the ordinal depth supervision (line~3), elevating the model to state-of-the-art results.

\begin{table}
\centering
\small
\hspace{-3mm}
\tabcolsep=1.0mm
\begin{tabular}{@{}l |lll |lll |l @{}}
\toprule
& \multicolumn{3}{c}{Walking} & \multicolumn{3}{c}{Jogging} &\\
& S1 & S3 & S3 & S1 & S2 & S3 & Avg\\
\midrule
Radwan~\etal~\cite{radwan2013monocular}   & 75.1 & 99.8 & 93.8 & 79.2 & 89.8 & 99.4 & 89.5\\
Wang~\etal~\cite{wang2014robust}          & 71.9 & 75.7 & 85.3 & 62.6 & 77.7 & 54.4 & 71.3\\
Simo-Serra~\etal~\cite{simo2013joint}     & 65.1 & 48.6 & 73.5 & 74.2 & 46.6 & 32.2 & 56.7\\
Bo~\etal~\cite{bo2010twin}                & 46.4 & 30.3 & 64.9 & 64.5 & 48.0 & 38.2 & 48.7\\
Kostrikov~\etal~\cite{kostrikov2014depth} & 44.0 & 30.9 & 41.7 & 57.2 & 35.0 & 33.3 & 40.3\\
Yasin~\etal~\cite{yasin2016dual}          & 35.8 & 32.4 & 41.6 & 46.6 & 41.4 & 35.4 & 38.9\\
Moreno-Noguer~\cite{moreno20163d}      & 19.7 & 13.0 & {\bf24.9} & 39.7 & 20.0 & 21.0 & 26.9\\
Pavlakos~\etal~\cite{pavlakos2016coarse}          & 22.1 & 21.9 & 29.0 & 29.8 & 23.6 & 26.0 & 25.5\\
Martinez~\etal~\cite{martinez2017simple}                      & 19.7 & 17.4 & 46.8 & 26.9 & 18.2 & 18.6 & 24.6\\
\midrule
Ours                      & {\bf18.8} & {\bf12.7} & 29.2 & {\bf23.5} & {\bf15.4} & {\bf14.5} & {\bf18.3}\\
\bottomrule
\end{tabular}
\vspace{-3pt}
\caption{Results on the HumanEva-I~\cite{sigal2010humaneva} dataset. Numbers are reconstruction errors (mm). The results of all approaches are obtained from the original papers.
}
\label{tab:humaneva}
\end{table}

\subsection{Comparison with state-of-the-art}\label{sec:sota}

\noindent
\textbf{Human3.6M}: 
We use for evaluation the same ConvNet with the previous section, which follows a mixed training strategy and includes the reconstruction component. The detailed results in terms of mean per joint error and reconstruction error are presented in Tables~\ref{tab:hm36m_mpjpe} and~\ref{tab:hm36m_rec} respectively. Our complete approach achieves state-of-the-art results across all actions and metrics, with relative error reduction over 10\% on average. Since most other works (e.g.,~\cite{tome2017lifting,zhou2017towards,tekin2017learning,martinez2017simple}) also use in-the-wild images with 2D keypoints for supervision, most of the improvement for our approach comes from augmenting training with ordinal depth relations for these examples. In particular, the error decrease with respect to previous work is more significant for challenging actions like Sitting Down, Photo or Sitting, with a lot of self-occlusions and rare poses. This benefit can be attributed to the greater variety of the LSP+MPII Ordinal images not just in terms of appearance (this also benefits the other approaches), but mainly in terms of 3D poses which are observed from our ConvNet in a weak 3D form.

\noindent
\textbf{HumanEva-I}: 
The ConvNet architecture remains the same, where HumanEva-I and LSP+MPII Ordinal images are used for mixed training. The reconstruction component is trained only on HumanEva-I MoCap. Our results are presented in Table~\ref{tab:humaneva} and show important accuracy benefit over previous approaches. On average, the relative error reduction is again over 10\%, which is a solid improvement considering the numbers for this dataset have mostly saturated.

\noindent
\textbf{MPI-INF-3DHP}:
For MPI-INF-3DHP, we report results using the same ConvNet we trained for Human3.6M, with Human3.6M and LSP+MPII Ordinal images. In Table~\ref{tab:mpi_inf_3dhp} we compare with two recent baselines which are not trained on this dataset, and we outperform them, with particularly large margin for the Outdoor sequence.

\begin{table}
\centering
\hspace{-3mm}
\tabcolsep=0.75mm
\begin{tabular}{@{}cccccc@{}}
\toprule
\multirow{3}{*}{Approach} & Studio & Studio & \multirow{2}{*}{Outdoor} & \multirow{2}{*}{All} & \multirow{2}{*}{All} \\
 & GS & no GS &  & &  \\
 \cmidrule{2-6}
 & 3DPCK & 3DPCK & 3DPCK & 3DPCK & AUC \\
\midrule
Mehta \etal \cite{mehta2017monocular} & 70.8 & 62.3 & 58.8 & 64.7 & 31.7 \\
Zhou \etal \cite{zhou2017towards} & 71.1 & {\bf 64.7} & 72.7 & 69.2 & 32.5 \\
\midrule
Ours & {\bf 76.5} & 63.1 & {\bf 77.5} & {\bf 71.9} & {\bf 35.3} \\
\bottomrule
\end{tabular}
\vspace{-3pt}
\caption{Detailed results on the test set of MPI-INF-3DHP~\cite{mehta2017monocular}. The results for all approaches are taken from the original papers. No training data from this dataset have been used for training by any method.}
\label{tab:mpi_inf_3dhp}
\end{table}

\subsection{Qualitative evaluation}

\begin{figure}[t]
\includegraphics[width=1\linewidth,trim={0cm 14cm 0cm 0cm},clip]{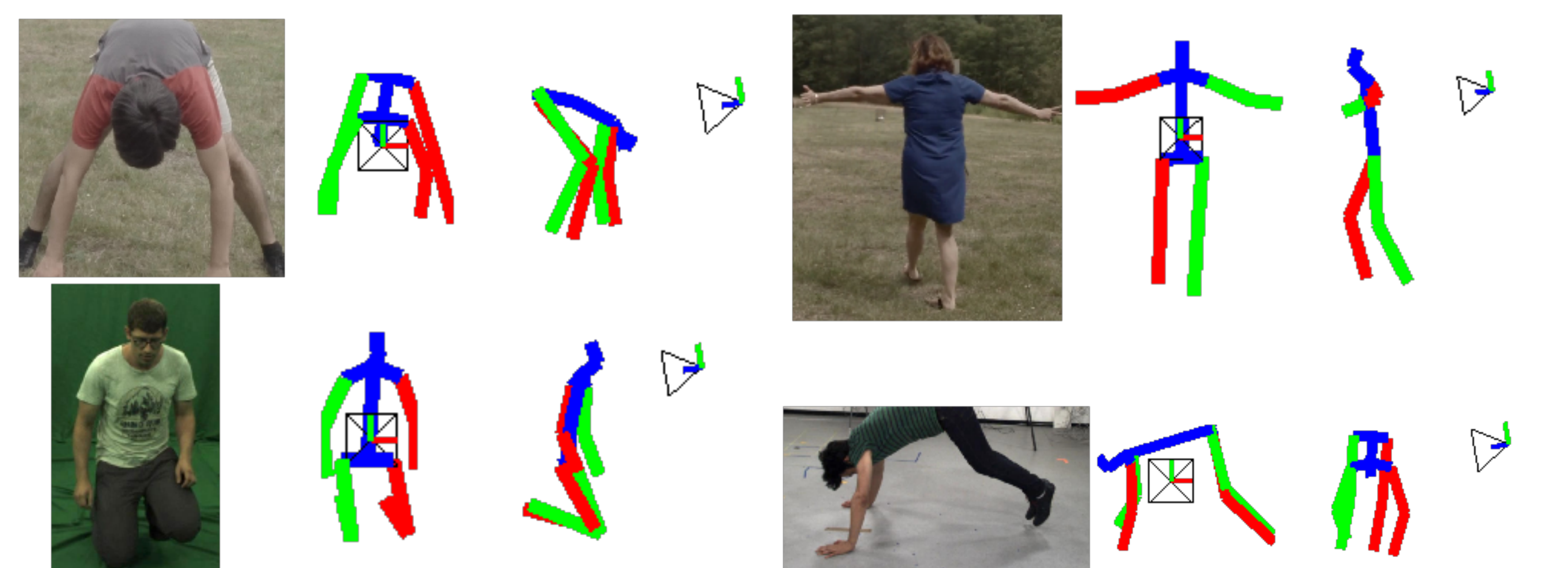}
\vspace{-10pt}
\caption{
Typical qualitative results from MPI-INF-3DHP, from the original and a novel viewpoint.
}
\label{fig:qualitative}
\end{figure}

In Figure~\ref{fig:qualitative} we have collected a sample of 3D pose output for our approach, focusing on MPI-INF-3DHP, since it is the main dataset that we evaluate without touching the training data. A richer collection of success and failure examples is included in the supplementary material.

\section{Summary}
The goal of this paper was to present a solution for training end-to-end ConvNets for 3D human pose estimation in the absence of accurate 3D ground truth, by using a weaker supervision signal in the form of ordinal depth relations of the joints. We investigated the flexibility of these ordinal relations by incorporating them in recent ConvNet architectures for 3D human pose and demonstrated competitive performance with their fully supervised versions. Furthermore, we extended the MPII and LSP datasets with ordinal depth annotations for the human joints, allowing us to present compelling results for non-studio conditions. Finally, these annotations were incorporated in the training procedure of recent ConvNets for 3D human pose, achieving state-of-the-art results in the standard benchmarks.

\vspace{1em}
\footnotesize
\noindent
{\bf Project Page:} \url{https://www.seas.upenn.edu/~pavlakos/projects/ordinal}

\vspace{0.5em}
\footnotesize
\noindent
{\bf Acknowledgements:} We gratefully appreciate support through the following grants: NSF-IIP-1439681 (I/UCRC), ARL RCTA W911NF-10-2-0016, ONR N00014-17-1-2093, DARPA FLA program and NSF/IUCRC.

{\small
\bibliographystyle{ieee}
\bibliography{egbib}

\begin{thebibliography}{10}\itemsep=-1pt

\bibitem{akhter2015pose}
I.~Akhter and M.~J. Black.
\newblock Pose-conditioned joint angle limits for 3{D} human pose
  reconstruction.
\newblock In {\em CVPR}, 2015.

\bibitem{andriluka2014mpii}
M.~Andriluka, L.~Pishchulin, P.~Gehler, and B.~Schiele.
\newblock {2D} human pose estimation: New benchmark and state of the art
  analysis.
\newblock In {\em CVPR}, 2014.

\bibitem{bell2014intrinsic}
S.~Bell, K.~Bala, and N.~Snavely.
\newblock Intrinsic images in the wild.
\newblock {\em ACM Transactions on Graphics (TOG)}, 33(4):159, 2014.

\bibitem{bo2010twin}
L.~Bo and C.~Sminchisescu.
\newblock Twin {G}aussian processes for structured prediction.
\newblock {\em IJCV}, 87(1-2):28--52, 2010.

\bibitem{bogo2016keep}
F.~Bogo, A.~Kanazawa, C.~Lassner, P.~Gehler, J.~Romero, and M.~J. Black.
\newblock Keep it {SMPL}: Automatic estimation of 3{D} human pose and shape
  from a single image.
\newblock In {\em ECCV}, 2016.

\bibitem{bourdev2009poselets}
L.~Bourdev and J.~Malik.
\newblock Poselets: Body part detectors trained using 3{D} human pose
  annotations.
\newblock In {\em ICCV}, 2009.

\bibitem{burges2005learning}
C.~Burges, T.~Shaked, E.~Renshaw, A.~Lazier, M.~Deeds, N.~Hamilton, and
  G.~Hullender.
\newblock Learning to rank using gradient descent.
\newblock In {\em ICML}, 2005.

\bibitem{cao2007learning}
Z.~Cao, T.~Qin, T.-Y. Liu, M.-F. Tsai, and H.~Li.
\newblock Learning to rank: from pairwise approach to listwise approach.
\newblock In {\em ICML}, 2007.

\bibitem{cao2016realtime}
Z.~Cao, T.~Simon, S.-E. Wei, and Y.~Sheikh.
\newblock Realtime multi-person 2{D} pose estimation using part affinity
  fields.
\newblock In {\em CVPR}, 2017.

\bibitem{chen20163d}
C.-H. Chen and D.~Ramanan.
\newblock 3{D} human pose estimation = 2{D} pose estimation + matching.
\newblock In {\em CVPR}, 2017.

\bibitem{chen2016attention}
L.-C. Chen, Y.~Yang, J.~Wang, W.~Xu, and A.~L. Yuille.
\newblock Attention to scale: Scale-aware semantic image segmentation.
\newblock In {\em CVPR}, 2016.

\bibitem{chen2016single}
W.~Chen, Z.~Fu, D.~Yang, and J.~Deng.
\newblock Single-image depth perception in the wild.
\newblock In {\em NIPS}, 2016.

\bibitem{chen2016synthesizing}
W.~Chen, H.~Wang, Y.~Li, H.~Su, Z.~Wang, C.~Tu, D.~Lischinski, D.~Cohen-Or, and
  B.~Chen.
\newblock Synthesizing training images for boosting human 3{D} pose estimation.
\newblock In {\em 3DV}, 2016.

\bibitem{du2016marker}
Y.~Du, Y.~Wong, Y.~Liu, F.~Han, Y.~Gui, Z.~Wang, M.~Kankanhalli, and W.~Geng.
\newblock Marker-less 3{D} human motion capture with monocular image sequence
  and height-maps.
\newblock In {\em ECCV}, 2016.

\bibitem{insafutdinov2016articulated}
E.~Insafutdinov, M.~Andriluka, L.~Pishchulin, S.~Tang, B.~Andres, and
  B.~Schiele.
\newblock Articulated multi-person tracking in the wild.
\newblock In {\em CVPR}, 2016.

\bibitem{ionescu2014human}
C.~Ionescu, D.~Papava, V.~Olaru, and C.~Sminchisescu.
\newblock Human3.6{M}: Large scale datasets and predictive methods for 3{D}
  human sensing in natural environments.
\newblock {\em PAMI}, 36(7):1325--1339, 2014.

\bibitem{jahangiri2017generating}
E.~Jahangiri and A.~L. Yuille.
\newblock Generating multiple hypotheses for human 3{D} pose consistent with
  2{D} joint detections.
\newblock In {\em ICCVW}, 2017.

\bibitem{johnson2010clustered}
S.~Johnson and M.~Everingham.
\newblock Clustered pose and nonlinear appearance models for human pose
  estimation.
\newblock In {\em BMVC}, 2010.

\bibitem{kazemi2013multi}
V.~Kazemi, M.~Burenius, H.~Azizpour, and J.~Sullivan.
\newblock Multi-view body part recognition with random forests.
\newblock In {\em BMVC}, 2013.

\bibitem{kostrikov2014depth}
I.~Kostrikov and J.~Gall.
\newblock Depth sweep regression forests for estimating 3{D} human pose from
  images.
\newblock In {\em BMVC}, 2014.

\bibitem{lassner2017unite}
C.~Lassner, J.~Romero, M.~Kiefel, F.~Bogo, M.~J. Black, and P.~V. Gehler.
\newblock Unite the people: Closing the loop between 3{D} and 2{D} human
  representations.
\newblock In {\em CVPR}, 2017.

\bibitem{li20143d}
S.~Li and A.~B. Chan.
\newblock 3{D} human pose estimation from monocular images with deep
  convolutional neural network.
\newblock In {\em ACCV}, 2014.

\bibitem{maji2011action}
S.~Maji, L.~Bourdev, and J.~Malik.
\newblock Action recognition from a distributed representation of pose and
  appearance.
\newblock In {\em CVPR}, 2011.

\bibitem{marinoiu2016pictorial}
E.~Marinoiu, D.~Papava, and C.~Sminchisescu.
\newblock Pictorial human spaces: A computational study on the human perception
  of 3{D} articulated poses.
\newblock {\em IJCV}, 119(2):194--215, 2016.

\bibitem{martinez2017simple}
J.~Martinez, R.~Hossain, J.~Romero, and J.~J. Little.
\newblock A simple yet effective baseline for 3{D} human pose estimation.
\newblock In {\em ICCV}, 2017.

\bibitem{mehta2017monocular}
D.~Mehta, H.~Rhodin, D.~Casas, O.~Sotnychenko, W.~Xu, and C.~Theobalt.
\newblock Monocular 3{D} human pose estimation in the wild using improved {CNN}
  supervision.
\newblock In {\em 3DV}, 2017.

\bibitem{mehta2017vnect}
D.~Mehta, S.~Sridhar, O.~Sotnychenko, H.~Rhodin, M.~Shafiei, H.-P. Seidel,
  W.~Xu, D.~Casas, and C.~Theobalt.
\newblock {VN}ect: Real-time 3{D} human pose estimation with a single {RGB}
  camera.
\newblock {\em ACM Transactions on Graphics}, 36, 2017.

\bibitem{moreno20163d}
F.~Moreno-Noguer.
\newblock 3{D} human pose estimation from a single image via distance matrix
  regression.
\newblock In {\em CVPR}, 2017.

\bibitem{narihira2015learning}
T.~Narihira, M.~Maire, and S.~X. Yu.
\newblock Learning lightness from human judgement on relative reflectance.
\newblock In {\em CVPR}, 2015.

\bibitem{newell2016associative}
A.~Newell, Z.~Huang, and J.~Deng.
\newblock Associative embedding: End-to-end learning for joint detection and
  grouping.
\newblock In {\em NIPS}, 2017.

\bibitem{newell2016stacked}
A.~Newell, K.~Yang, and J.~Deng.
\newblock Stacked hourglass networks for human pose estimation.
\newblock In {\em ECCV}, 2016.

\bibitem{pavlakos2016coarse}
G.~Pavlakos, X.~Zhou, K.~G. Derpanis, and K.~Daniilidis.
\newblock Coarse-to-fine volumetric prediction for single-image 3{D} human
  pose.
\newblock In {\em CVPR}, 2017.

\bibitem{pfister2015flowing}
T.~Pfister, J.~Charles, and A.~Zisserman.
\newblock Flowing convnets for human pose estimation in videos.
\newblock In {\em ICCV}, 2015.

\bibitem{pishchulin2016deepcut}
L.~Pishchulin, E.~Insafutdinov, S.~Tang, B.~Andres, M.~Andriluka, P.~V. Gehler,
  and B.~Schiele.
\newblock Deep{C}ut: Joint subset partition and labeling for multi person pose
  estimation.
\newblock In {\em CVPR}, 2016.

\bibitem{pons2014posebits}
G.~Pons-Moll, D.~J. Fleet, and B.~Rosenhahn.
\newblock Posebits for monocular human pose estimation.
\newblock In {\em CVPR}, 2014.

\bibitem{popa2017deep}
A.-I. Popa, M.~Zanfir, and C.~Sminchisescu.
\newblock Deep multitask architecture for integrated 2{D} and 3{D} human
  sensing.
\newblock In {\em CVPR}, 2017.

\bibitem{radwan2013monocular}
I.~Radwan, A.~Dhall, and R.~Goecke.
\newblock Monocular image 3{D} human pose estimation under self-occlusion.
\newblock In {\em ICCV}, 2013.

\bibitem{ramakrishna2012}
V.~Ramakrishna, T.~Kanade, and Y.~Sheikh.
\newblock Reconstructing 3{D} human pose from 2{D} image landmarks.
\newblock In {\em ECCV}, 2012.

\bibitem{rogez2016mocap}
G.~Rogez and C.~Schmid.
\newblock Mo{C}ap-guided data augmentation for {3D} pose estimation in the
  wild.
\newblock In {\em NIPS}, 2016.

\bibitem{rogez2017lcr}
G.~Rogez, P.~Weinzaepfel, and C.~Schmid.
\newblock {LCR-N}et: Localization-classification-regression for human pose.
\newblock In {\em CVPR}, 2017.

\bibitem{sarafianos20163d}
N.~Sarafianos, B.~Boteanu, B.~Ionescu, and I.~A. Kakadiaris.
\newblock 3{D} human pose estimation: A review of the literature and analysis
  of covariates.
\newblock {\em CVIU}, 152:1--20, 2016.

\bibitem{sigal2010humaneva}
L.~Sigal, A.~O. Balan, and M.~J. Black.
\newblock {HumanEva}: Synchronized video and motion capture dataset and
  baseline algorithm for evaluation of articulated human motion.
\newblock {\em IJCV}, 87(1-2):4--27, 2010.

\bibitem{simo2013joint}
E.~Simo-Serra, A.~Quattoni, C.~Torras, and F.~Moreno-Noguer.
\newblock A joint model for 2{D} and 3{D} pose estimation from a single image.
\newblock In {\em CVPR}, 2013.

\bibitem{sun2017compositional}
X.~Sun, J.~Shang, S.~Liang, and Y.~Wei.
\newblock Compositional human pose regression.
\newblock In {\em ICCV}, 2017.

\bibitem{taylor2000reconstruction}
C.~J. Taylor.
\newblock Reconstruction of articulated objects from point correspondences in a
  single uncalibrated image.
\newblock In {\em CVPR}, 2000.

\bibitem{taylor2008softrank}
M.~Taylor, J.~Guiver, S.~Robertson, and T.~Minka.
\newblock Soft{R}ank: optimizing non-smooth rank metrics.
\newblock In {\em Proceedings of the 2008 International Conference on Web
  Search and Data Mining}, pages 77--86. ACM, 2008.

\bibitem{tekin2016structured}
B.~Tekin, I.~Katircioglu, M.~Salzmann, V.~Lepetit, and P.~Fua.
\newblock Structured prediction of {3D} human pose with deep neural networks.
\newblock In {\em BMVC}, 2016.

\bibitem{tekin2017learning}
B.~Tekin, P.~Marquez~Neila, M.~Salzmann, and P.~Fua.
\newblock Learning to fuse 2{D} and 3{D} image cues for monocular body pose
  estimation.
\newblock In {\em ICCV}, 2017.

\bibitem{tekin2015direct}
B.~Tekin, A.~Rozantsev, V.~Lepetit, and P.~Fua.
\newblock Direct prediction of {3D} body poses from motion compensated
  sequences.
\newblock In {\em CVPR}, 2016.

\bibitem{todd2003visual}
J.~T. Todd and J.~F. Norman.
\newblock The visual perception of 3-{D} shape from multiple cues: Are
  observers capable of perceiving metric structure?
\newblock {\em Perception \& Psychophysics}, 65(1):31--47, 2003.

\bibitem{tome2017lifting}
D.~Tome, C.~Russell, and L.~Agapito.
\newblock Lifting from the deep: Convolutional 3{D} pose estimation from a
  single image.
\newblock In {\em CVPR}, 2017.

\bibitem{tompson2014joint}
J.~J. Tompson, A.~Jain, Y.~LeCun, and C.~Bregler.
\newblock Joint training of a convolutional network and a graphical model for
  human pose estimation.
\newblock In {\em NIPS}, 2014.

\bibitem{varol2017learning}
G.~Varol, J.~Romero, X.~Martin, N.~Mahmood, M.~Black, I.~Laptev, and C.~Schmid.
\newblock Learning from synthetic humans.
\newblock In {\em CVPR}, 2017.

\bibitem{wang2014robust}
C.~Wang, Y.~Wang, Z.~Lin, A.~L. Yuille, and W.~Gao.
\newblock Robust estimation of 3{D} human poses from a single image.
\newblock In {\em CVPR}, 2014.

\bibitem{wei2016cpm}
S.-E. Wei, V.~Ramakrishna, T.~Kanade, and Y.~Sheikh.
\newblock Convolutional pose machines.
\newblock In {\em CVPR}, 2016.

\bibitem{wu2016single}
J.~Wu, T.~Xue, J.~J. Lim, Y.~Tian, J.~B. Tenenbaum, A.~Torralba, and W.~T.
  Freeman.
\newblock Single image {3D} interpreter network.
\newblock In {\em ECCV}, 2016.

\bibitem{xia2016zoom}
F.~Xia, P.~Wang, L.-C. Chen, and A.~L. Yuille.
\newblock Zoom better to see clearer: Human and object parsing with
  hierarchical auto-zoom net.
\newblock In {\em ECCV}, 2016.

\bibitem{xiang2016objectnet3d}
Y.~Xiang, W.~Kim, W.~Chen, J.~Ji, C.~Choy, H.~Su, R.~Mottaghi, L.~Guibas, and
  S.~Savarese.
\newblock Object{N}et3{D}: A large scale database for 3{D} object recognition.
\newblock In {\em ECCV}, 2016.

\bibitem{xiang2014beyond}
Y.~Xiang, R.~Mottaghi, and S.~Savarese.
\newblock Beyond {PASCAL}: A benchmark for 3{D} object detection in the wild.
\newblock In {\em WACV}, 2014.

\bibitem{xiaohan2017monocular}
B.~Xiaohan~Nie, P.~Wei, and S.-C. Zhu.
\newblock Monocular 3{D} human pose estimation by predicting depth on joints.
\newblock In {\em ICCV}, 2017.

\bibitem{yang2017learning}
W.~Yang, S.~Li, W.~Ouyang, H.~Li, and X.~Wang.
\newblock Learning feature pyramids for human pose estimation.
\newblock In {\em ICCV}, 2017.

\bibitem{yasin2016dual}
H.~Yasin, U.~Iqbal, B.~Kr{\"u}ger, A.~Weber, and J.~Gall.
\newblock A dual-source approach for 3{D} pose estimation from a single image.
\newblock In {\em CVPR}, 2016.

\bibitem{zhou2015learning}
T.~Zhou, P.~Kr\"ahenb\"uhl, and A.~A. Efros.
\newblock Learning data-driven reflectance priors for intrinsic image
  decomposition.
\newblock In {\em ICCV}, 2015.

\bibitem{zhou2017towards}
X.~Zhou, Q.~Huang, X.~Sun, X.~Xue, and Y.~Wei.
\newblock Towards 3{D} human pose estimation in the wild: A weakly-supervised
  approach.
\newblock In {\em ICCV}, 2017.

\bibitem{zhou2016deep}
X.~Zhou, X.~Sun, W.~Zhang, S.~Liang, and Y.~Wei.
\newblock Deep kinematic pose regression.
\newblock In {\em ECCVW}, 2016.

\bibitem{zhou2016sparse}
X.~Zhou, M.~Zhu, S.~Leonardos, and K.~Daniilidis.
\newblock Sparse representation for 3{D} shape estimation: A convex relaxation
  approach.
\newblock {\em PAMI}, 2016.

\bibitem{zhou2016sparseness}
X.~Zhou, M.~Zhu, S.~Leonardos, K.~Derpanis, and K.~Daniilidis.
\newblock Sparseness meets deepness: {3D} human pose estimation from monocular
  video.
\newblock In {\em CVPR}, 2016.

\bibitem{zhou2017monocap}
X.~Zhou, M.~Zhu, G.~Pavlakos, S.~Leonardos, K.~G. Derpanis, and K.~Daniilidis.
\newblock Mono{C}ap: Monocular human motion capture using a {CNN} coupled with
  a geometric prior.
\newblock {\em PAMI}, 2018.

\bibitem{zoran2015learning}
D.~Zoran, P.~Isola, D.~Krishnan, and W.~T. Freeman.
\newblock Learning ordinal relationships for mid-level vision.
\newblock In {\em ICCV}, 2015.

\end{thebibliography}
}

\end{document}